\documentclass{article}

\usepackage{arxiv}

\usepackage[utf8]{inputenc} 
\usepackage[T1]{fontenc}    
\usepackage[hidelinks,pagebackref]{hyperref}       
\usepackage{url}            
\usepackage{booktabs}       
\usepackage{amsfonts}       
\usepackage{microtype}      
\usepackage{graphicx}
\usepackage[round]{natbib}
\usepackage{doi}
\usepackage{float}
\usepackage{subcaption}
\usepackage{xcolor}

\definecolor{linkblue}{RGB}{20,20,180}
\hypersetup{pdfborder={0 0 0}}   

\newcommand{\resline}[3]{%
  \noindent\raisebox{-0.15\height}{#1}\hspace{0.5em}%
  \textbf{\href{#3}{\textcolor{linkblue}{#2}}}\par\vspace{0.4em}%
}

\usepackage[english]{babel}

\title{OpenPVMapper: A Multi-source, Nationwide Database of Rooftop Photovoltaic Systems in France}

\author{%
\href{https://orcid.org/0000-0002-7774-4302}{\includegraphics[scale=0.06]{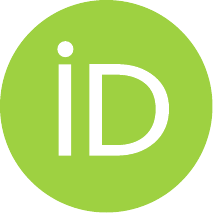}}\hspace{1mm}Gabriel Kasmi$^{1}$\\ \quad
$^1$\texttt{gabriel.kasmi@deeppvmapper.fr}\\
}

\date{}

\renewcommand{\headeright}{}
\renewcommand{\undertitle}{Version: August 2026}
\renewcommand{\shorttitle}{OpenPVMapper}

\begin{document}
\maketitle

\begin{abstract}
Rooftop photovoltaic (PV) systems account for the vast majority of PV grid connections, yet no open, comprehensive, installation-level dataset of these systems exists: public registries aggregate data only above a capacity threshold, and remote sensing-based detection efforts, while extensive, are typically confined to a single method, a limited geographic scope, or a single point in time. We introduce OpenPVMapper, a nationwide, multi-source database of rooftop PV installations in mainland France, built by aggregating and reconciling complementary sources: a deep learning-based detection pipeline deployed on nationwide aerial imagery, OpenStreetMap and a probabilistic building-level detection dataset. The resulting database contains 1,135,850 installations, totaling approximately 15.01~GWp of installed capacity, each documented with its provenance, detection method, and, where available, a manual validation flag. Manual review of a stratified sample of 1,862 installations places the database's overall precision at approximately 74--75\%, with corroboration across independent sources bringing a substantial, quantified precision gain. By aggregating independent sources rather than relying on any single detection method, OpenPVMapper reaches a level of confidence beyond what any one source could provide on its own, while remaining extensible to further sources as they become available. It is released under an open (CC-BY) license alongside the full source code used to build it.
\begin{center}
\vspace{0.5em}
\resline{\includegraphics[height=1.05em]{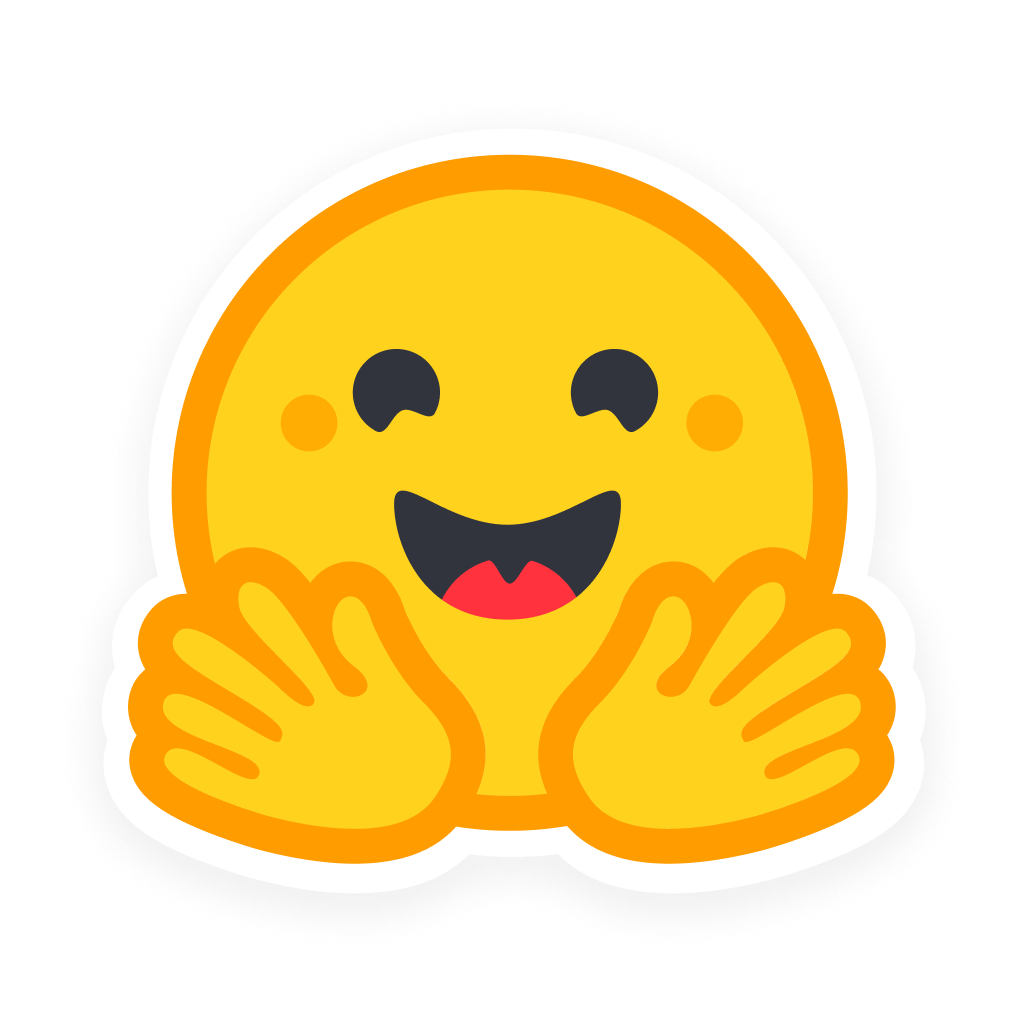}}%
  {Dataset repository}{https://huggingface.co/datasets/gabrielkasmi/openpvmapper}
\end{center}
\end{abstract}

\section*{Background \& Summary}

Rooftop photovoltaic (PV) systems represent the vast majority of PV
installations connected to the electricity grid, and their
number keeps growing at a fast, highly decentralized pace: new systems
are installed daily by individual households and businesses,
largely independently of one another and often outside any centralized
planning process. This decentralization is precisely what makes rooftop
PV valuable for the energy transition, but it also makes it uniquely
difficult to monitor. Where public administrative registries exist, they typically report only
aggregated capacity above a fixed threshold (e.g., 36~kWp in France),
mask fine-grained location for smaller installations (below 30~kWp in
Germany's Marktstammdatenregister \citep{kotthoff_monitoring_2024}), or
can otherwise be incomplete or outdated \citep{kasmi_enhancing_2024}.

Remote sensing-based detection has emerged as a promising way to
fill this gap
\citep{malof_automatic_2016,yu_deepsolar_2018,mayer_deepsolar_2020,kausika_geoai_2021,mayer_3d-pv-locator_2022,lindahl_mapping_2023,thebault_comprehensive_2025,kasmi_enhancing_2024},
and a growing number of studies have demonstrated that rooftop PV systems
can be automatically detected and characterized from aerial or satellite
imagery at scale. However, any single detection effort, however accurate,
is bound by its own scope: a given detection method covers a fixed
period, a fixed geographic extent, and inherits its own detection biases
\citep{kasmi_space-scale_2025}. 

Community-contributed geographic data offers high per-record quality --
installations are verified by human mappers -- but is scarcely and
unevenly available: building a comprehensive crowdsourced PV layer from
scratch, as demonstrated for the UK by \citet{stowell_harmonised_2020},
requires a dedicated, sustained community effort that is not available in
every country or region, and, at a global scale, PV is simply one asset
class among many that a general-purpose crowdsourced map like OSM cannot
be expected to cover exhaustively on its own (millions of individual
rooftop systems worldwide is a lot of manual mapping to ask of any single
community).

The core strength of crowdsourcing, however, is not any single
contribution but the redundancy across independent ones: aggregating
multiple noisy, independent labels reliably outperforms any single
label, expert or not \citep{dawid_maximum_1979,snow_cheap_2008,sheng_get_2008},
a principle that has produced high-quality datasets across domains, from
species observation (iNaturalist \citep{van_horn_inaturalist_2018}, eBird \citep{sullivan_ebird_2009}) to citizen science image
classification (Galaxy Zoo, \citep{lintott_galaxy_2008}), and, for rooftop PV specifically, BDAPPV
\citep{kasmi_crowdsourced_2023}. Volunteered geographic data such as OSM
inherits the same property whenever contributions are numerous enough to
be cross-checked \citep{haklay_how_2010,girres_quality_2010}, and this is
the principle OpenPVMapper applies to construct a reference database of
rooftop PV installations in France: rather than relying on a single
detection method, it cross-checks two independent, large-scale mapping
efforts, DeepPVMapper \citep{kasmi_enhancing_2024} and FRPV
\citep{thebault_comprehensive_2025}, to build an installation-level
database that is as extensive and as reliable as the two combined can
make it. OpenStreetMap is integrated as a third, smaller but highly
reliable source: its overall contribution is modest in scale, but wherever
present, human verification makes it the most trustworthy of the three
(see \emph{Technical Validation}).

Our central result is that automated detection methods act as prolific
but comparatively unreliable contributors -- their precision, taken alone,
is as low as 71\% -- while agreement between two independent methods
lifts precision to 97\%, approaching that of human-verified OpenStreetMap
annotations (98\%). This confirms, at the scale of an entire country,
that combining independent sources substantially improves detection
quality: the number of corroborating sources for a given installation can
itself be used as a proxy for its precision.

OpenPVMapper is, to our knowledge, the most comprehensive effort to date
to map and characterize individual rooftop PV systems in France. Totalling
1.1 million systems and approximately 15.0~GWp of installed capacity, it
covers more than two-thirds of the connected capacity registered by grid
operators (as of Q2 2025) and nearly all rooftop PV systems below
36~kWp (97\%) -- the power class least documented by existing public
data. The dataset is available at \url{https://zenodo.org/records/21534856}.

\begin{figure}[h]
    \centering
    \includegraphics[width=0.6\linewidth]{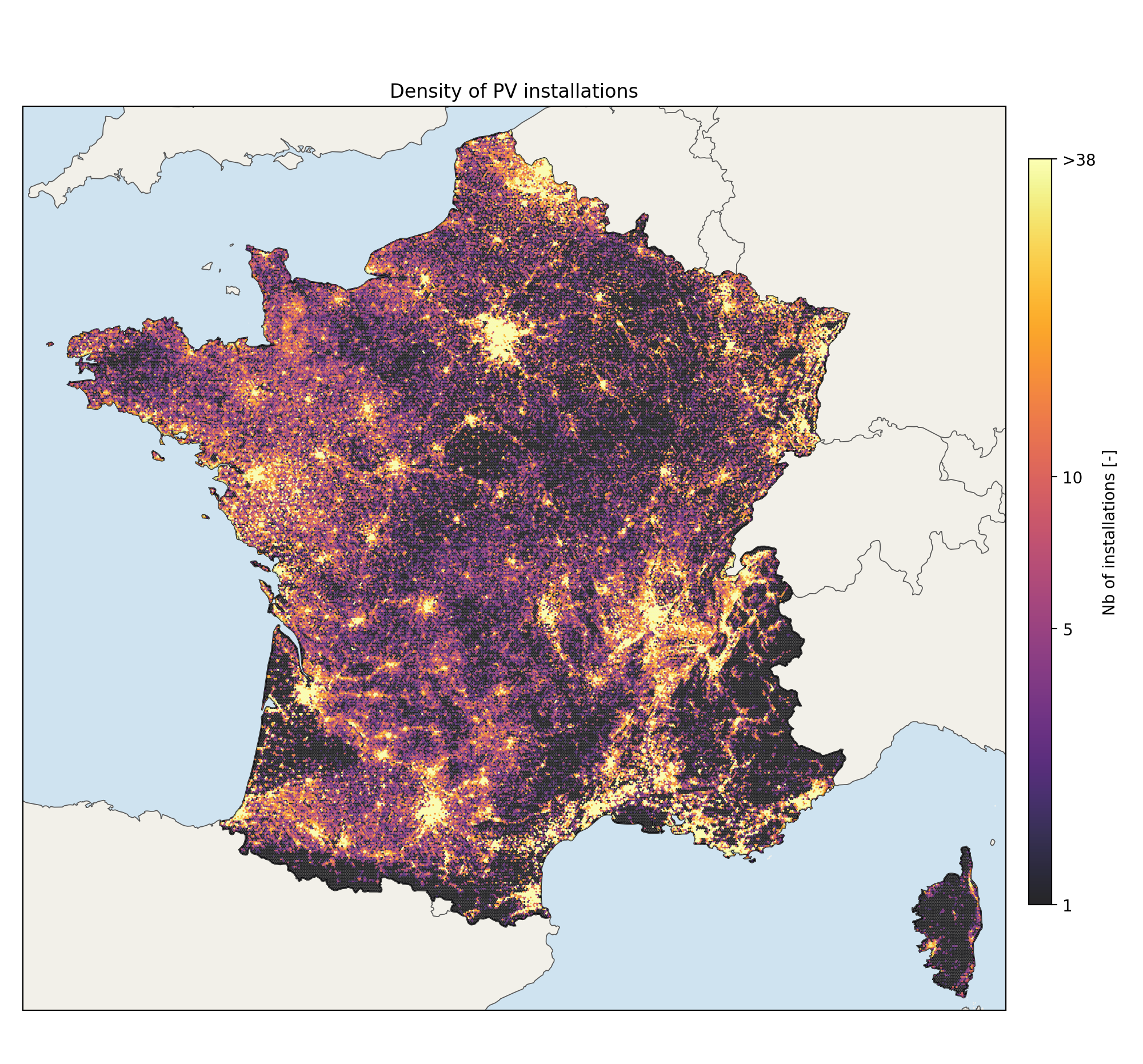}
    \caption{\textbf{Rooftop PV density recorded in OpenPVMapper.} Number of installations per hexagonal grid cell (mean area $\approx$~8~km$^2$) across mainland France, aggregated from the full, multi-source database (1,135,850 installations, cf.\ \emph{Data Overview}), on a logarithmic color scale to capture the wide range between sparsely covered rural areas and dense urban and peri-urban clusters. Beyond illustrating the finished product, the map reflects the combined footprint of all three sources rather than any single one: high-density areas are not artifacts of a single detector's coverage bias, since they persist even where DeepPVMapper, FRPV, and OpenStreetMap each contribute unevenly (\autoref{fig:cartogram}).}
    \label{fig:pv_density}
\end{figure}

\section*{Methods}

\subsection*{Nationwide mapping using DeepPVMapper}

The base layer of the database is produced by DeepPVMapper,
a two-stage deep learning pipeline deployed independently
on each of the 96 departments of mainland France, using
the IGN BD ORTHO aerial imagery (20~cm/pixel resolution).
Candidate image tiles are first screened by an
Inception-v3 classification model \citep{szegedy_rethinking_2016},
and positive tiles are then passed to a DeepLab-v3
segmentation model \citep{chen_deeplab_2018} to extract pixel-level
PV masks, which are converted into geolocalized polygons.
Each polygon is then characterized (surface, tilt, azimuth,
installed capacity) using PyPVRoof \citep{tremenbert_pypvroof_2023}.
On held-out test data, this pipeline reaches a classification
F1-score of 0.84 and a segmentation IoU of 0.86, and installed
capacity is estimated with a RMSE of 0.69~kWp
(cf.\ \citep{kasmi_towards_2022,kasmi_enhancing_2024} for a
comprehensive evaluation). Installations larger than 36~kWp or
smaller than a single PV module (300~Wp) are filtered out,
matching the residential and small commercial segment that
public registries do not cover.

Taken on its own, this pipeline inherits DeepPVMapper's own detection
biases -- extensively documented in \cite{kasmi_space-scale_2025} -- and its 36~kWp upper bound. Both limitations are lifted at
the database level: OpenStreetMap and other sources are not
bound by the same capacity threshold and can contribute larger
installations, and reconciling multiple independent detection
methods over the same territory reduces the false negative
rate of any single one of them (see \emph{Data Fusion and Enrichment}, below).

\subsection*{Multi-source acquisition}

\begin{figure}[h]
    \centering
    \includegraphics[width=\linewidth]{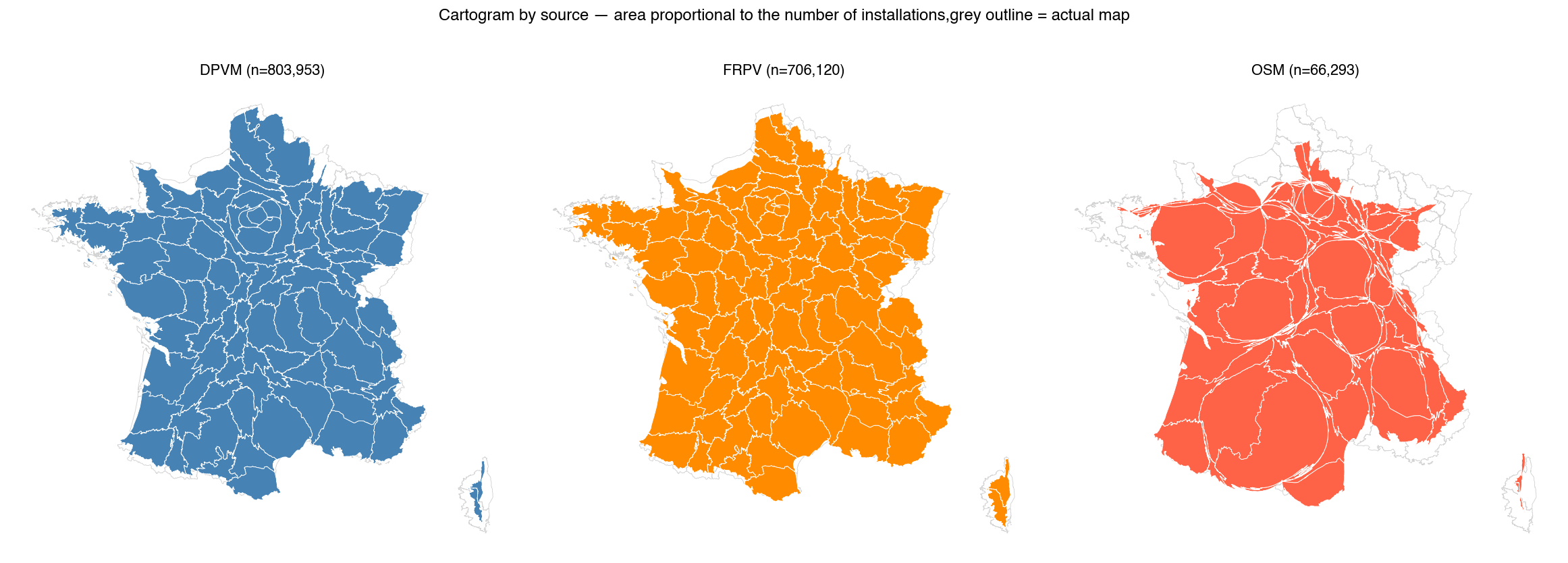}
    \caption{\textbf{Geographic concentration of installations, by source.} Each department's surface area is distorted proportionally to its number of installations for DeepPVMapper (DPVM), FRPV, and OpenStreetMap (OSM) respectively; the light grey outline shows the true map for reference. FRPV's coverage is close to uniform across mainland France, DPVM is only slightly more concentrated, while OSM's community-contributed coverage is markedly skewed toward a small number of departments.}
    \label{fig:cartogram}
\end{figure}

The three sources combined at this stage differ markedly in their
geographic footprint (\autoref{fig:cartogram}). Combining them is
therefore not just a matter of adding coverage: OSM's contribution is
concentrated in specific regions -- notably the south-west, where its
per-installation precision is also the highest of the three sources
(cf.\ \emph{Technical Validation}) -- while DeepPVMapper and FRPV provide
a comparatively even geographic baseline everywhere else. Aggregating the
three lets each source's local strength compensate for the others'
blind spots.

\paragraph{OpenStreetMap Data}
OpenStreetMap contributors manually map PV installations as
points or building-level polygons, tagged with an installation type.
Because this information is contributed and checked by human
mappers, it is treated as the most reliable geometry whenever
it is available (see \emph{Data Fusion and Enrichment}), rather than as just
another detection to reconcile with equal weight. OSM data
is retrieved from a national OSM extract, filtered on
PV-related tags, and split by department for processing.

\paragraph{Presence of PV on French Rooftops (FRPV)}
FRPV \citep{thebault_comprehensive_2025} is a building-wise rooftop
PV detection dataset covering 95 of the 96 French departments
at the time of this release (dataset in progressive construction).
Unlike DeepPVMapper or OSM, FRPV does not provide a
ready-to-use polygon: it reports, for each cadastral parcel,
a probability of hosting a PV installation, which requires
calibrating a decision threshold before it can be used as a
detection source (see \emph{Data Fusion and Enrichment}).

\paragraph{Validation data}
Two independent, manually annotated samples are used for quality
control rather than as detection sources: a precision sample of 18,337 points, each an existing detection manually confirmed as a true or false positive (4,371 confirmed false positives), and a recall sample of 13,366 points, each an independently sourced candidate installation (from OSM or manual annotation) manually checked against the detection database to identify installations missed by it. Together, these two campaigns represent 31,703 manually reviewed points, both drawn from an ongoing companion study.
For the recall sample specifically, a candidate is considered already present in the database if an existing detection lies within 30~m (7,916 of the 13,366 points); the remaining candidates are either confirmed false negatives (promoted as new installations, see \emph{Data Fusion and Enrichment}) or left aside when the imagery date makes their status ambiguous.

\paragraph{Future updates} Three streams of updates may be considered for future releases: updates to include additional mapping efforts as additional works provide mapping of PV systems over France, additional validation data, either coming from manual verification and/or retrieval from OSM, and releases due to the deployment of updated and more accurate mapping models. The pipeline's source-hierarchy design (see \emph{Data
Fusion and Enrichment}) is what makes this possible: adding a source
means slotting it into the existing hierarchy and matching logic, not
redesigning the schema.

\subsection*{Data Fusion and Enrichment}\label{sec:fusion}

Combining heterogeneous sources into a single, coherent database serves three objectives. First, \emph{uniformization}: each source has its own native schema, which is mapped onto a common target schema, using DeepPVMapper's own schema as the reference (see \emph{Data Record}, below). Second, \emph{enrichment}: combining sources adds information that no single source carries on its own -- successive DPVM imagery vintages document when an installation was first and last observed, agreement between independent methods increases confidence in a detection, and the two manually annotated samples act as a gold standard wherever they overlap with the database. Third, \emph{hierarchization}: when two sources provide a geometry for the same real-world installation, one polygon has to be retained as the reference geometry, following a fixed hierarchy of sources (correction $>$ OpenStreetMap $>$ DeepPVMapper $>$ third-party detections; FRPV is never used as a geometry source, cf.\ next paragraph).

\paragraph{Multi-vintage DPVM}

Most departments have between two and four successive DeepPVMapper
detection vintages available. The full set of raw, per-vintage detections
underlying this multi-vintage matching is released separately as
\texttt{raw.zip} (see \emph{Data Record}), so that users needing the
un-merged detection history rather than the resolved database can access it
directly.

Matching polygons across vintages (same spatial matching rule as across sources)
yields, for each installation, a \texttt{first\_seen} (oldest vintage where the installation is detected)
and \texttt{last\_seen} (most recent vintage) date. An installation observed in an early
vintage and absent from a later one is not removed from the database: it is retained
with its original \texttt{first\_seen}/\texttt{last\_seen} span, since a single missed vintage does
not imply the installation was decommissioned.

\paragraph{OpenStreetMap}
OSM installations are provided as polygons and,
being manually contributed and checked, are treated as the most reliable
available geometry: whenever an OSM polygon matches an existing
detection, it supersedes it as the reference geometry, and characteristics (surface, tilt, azimuth, installed capacity) are recomputed on this new polygon.

OSM is not designed, and does not aim, to exhaustively map every PV
installation at a global scale: with millions of individual rooftop
systems worldwide, comprehensive manual coverage of a single asset class
like rooftop PV is not a realistic goal for a general-purpose
crowdsourced map, and OSM's own PV coverage in France reflects this --
partial, geographically uneven, but wherever present, highly reliable
(\autoref{fig:cartogram}). This is precisely the practical case for
automated efforts such as OpenPVMapper: rather than substituting for OSM,
the pipeline is designed to be \emph{interoperable} with it (adopting its
tags and schema directly, as above) and to operate \emph{independently}
of any single community's mapping effort, so that the two can extend
each other -- automated detection filling OSM's geographic gaps, and OSM
supplying a trusted geometry wherever a human mapper has already
contributed one.

\paragraph{FRPV} As FRPV reports a per-building probability rather than a
ready-to-use detection, a decision threshold has to be calibrated
to decide which parcels are promoted as new installation candidates. This threshold was calibrated on a stratified sample of approximately 1,200 installations, sampled across probability bands to cover the full range of the score distribution, manually labelled blind to the underlying score, and evaluated with a precision-recall curve with bootstrap confidence intervals (\autoref{fig:frpv-threshold}, detailed in \emph{Technical Validation}). Parcels above the calibrated threshold that are not already matched to an existing detection are promoted as new installation candidates (points, without a polygon -- see \emph{Uniformization}, below).

\subsection*{Uniformization}\label{sec:uniformization}

The final step ensures that every installation in the database, regardless of its originating source, carries the same set of fields, populated as completely as possible (see \emph{Data Record}). Installations whose reference geometry was introduced or superseded by OpenStreetMap are re-characterized (surface, tilt, azimuth, installed capacity recomputed on the new polygon). Installations known only as a point (FRPV candidates above threshold, confirmed false negatives from the recall sample) have no usable geometry yet, so their coordinates is used to extract an image on which the same segmentation and characterization branch as DeepPVMapper's own pipeline.

\begin{figure}[h]
    \centering
    \includegraphics[width=\linewidth]{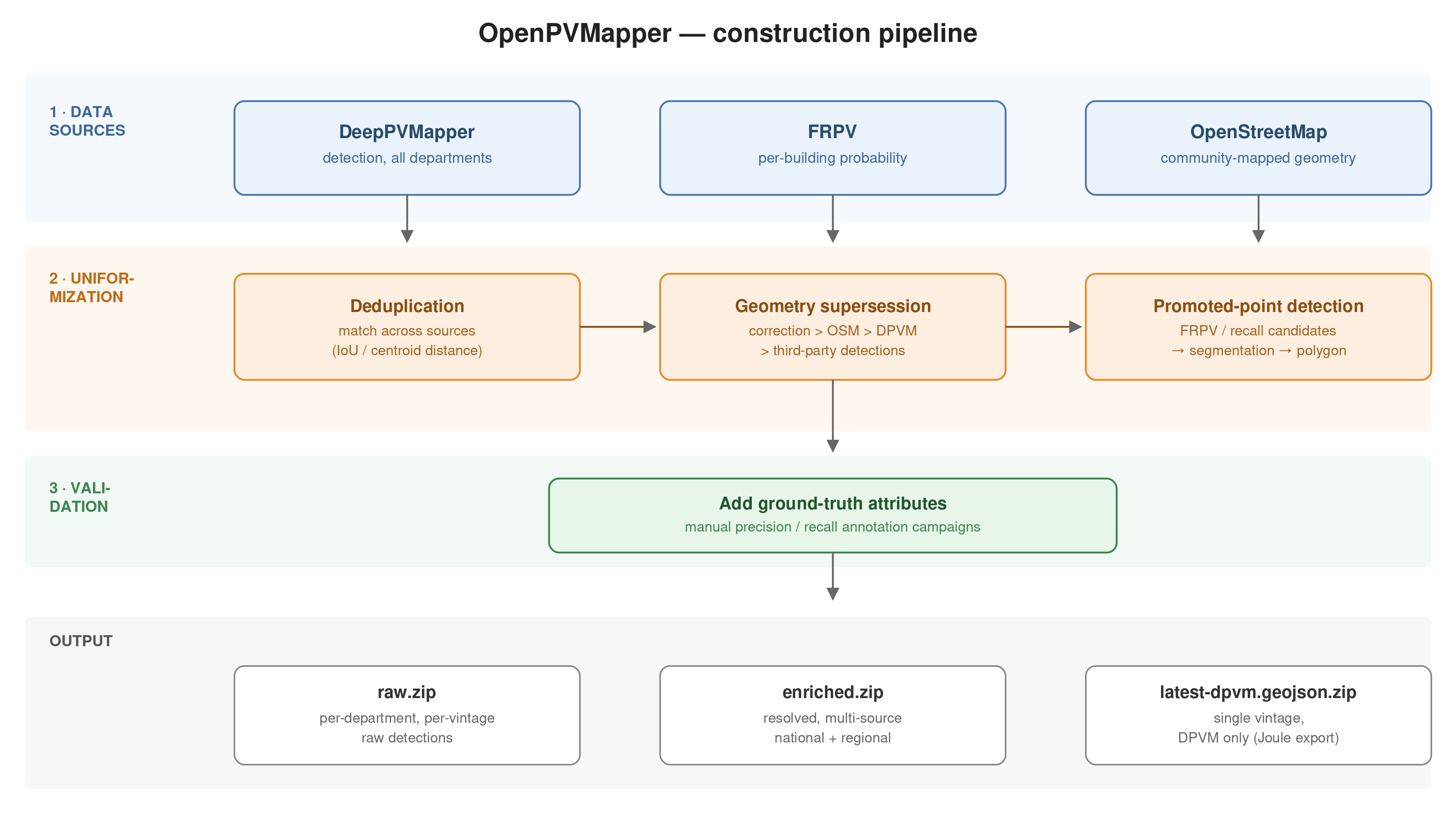}
    \caption{\textbf{Overview of the OpenPVMapper construction pipeline.} Detections from the three data sources (DeepPVMapper, FRPV, OpenStreetMap) are combined through a three-stage process: deduplication and matching across sources, geometry supersession following the fixed source hierarchy described in \emph{Data Fusion and Enrichment}, and detection/segmentation of FRPV and recall-promoted point candidates. Ground-truth attributes from the manual precision and recall annotation campaigns are added last. The pipeline produces three output archives: \texttt{raw.zip} (per-department, per-vintage raw detections), \texttt{enriched.zip} (the resolved, multi-source national database), and \texttt{latest-dpvm.geojson.zip} (a single-vintage, DeepPVMapper-only export).}
    \label{fig:pipeline}
\end{figure}

\autoref{fig:pipeline} depicts the overall data acquisition, uniformization and enrichment phrases to construct the database. 

\section*{Data Record}\label{sec:data-record}

The database is released on Zenodo (\texttt{DOI:}~\url{https://doi.org/10.5281/zenodo.21534856}) under an open license (CC-BY), and versioned: future releases update the same record rather than creating a new one. The dataset is also available on HuggingFace at the following URL: \url{https://huggingface.co/datasets/gabrielkasmi/openpvmapper}.

\paragraph{Files} The release is organized as three archives:
\begin{itemize}
    \item \texttt{enriched.zip}: the resolved, multi-source national database. Contains \texttt{enriched/enriched-national.geojson} (one polygon per installation, 1,135,850 rows, see \emph{Data Overview}) and \texttt{enriched/regions/\{region\}.geojson}, the same database split into the 13 mainland French administrative regions, for users who only need one region without downloading the full national file.
    \item \texttt{raw.zip}: the raw, per-department, per-vintage DeepPVMapper detections underlying the enriched database (\texttt{raw/D0xx\_20xx/arrays\_characteristics\_\{dept\}\_\{year\}.geojson}), together with \texttt{raw/index.json}, an index of the department/vintage combinations available.
    \item \texttt{latest-dpvm.geojson.zip}: a simplified, single-vintage, DPVM-only export (one row per installation on its most recent detected DeepPVMapper vintage, no multi-source merge) -- the export used in the companion Joule paper, 593,671 rows.
\end{itemize}
Duplicate-merge audit logs (which record was dropped, which record it was merged into during deduplication) are included alongside the enriched export so that no merge happens silently.

\paragraph{Fields} \autoref{tab:schema} describes the fields of the national database (\texttt{enriched-national.geojson} and the regional exports share the same schema). All files use \textbf{EPSG:4326 (WGS84)}.

\begin{table}[h]
\centering
\caption{Fields of the national database.}\label{tab:schema}
\begin{tabular}{l l p{8.3cm}}
\toprule
\textbf{Field} & \textbf{Type} & \textbf{Description} \\
\midrule
\texttt{array\_id} & string & Unique installation identifier. \\
\texttt{geometry} & polygon & Installation footprint, in the source selected by the geometry hierarchy (see \emph{Data Fusion and Enrichment}). \\
\texttt{insee} & string & INSEE code of the commune the installation is located in. \\
\texttt{dpt} & string & Department code (2 characters, \texttt{2A}/\texttt{2B} for Corsica, 3 characters for overseas departments). \\
\texttt{rnb\_id} & string, nullable & Identifier of the matching building in the French national building registry (RNB), when available. \\
\texttt{surface} & float32 & Estimated installation surface area, in m$^2$ (rounded to 2 decimals). \\
\texttt{tilt} & Int64, nullable & Estimated tilt angle, in degrees. \\
\texttt{azimuth} & Int64, nullable & Estimated azimuth angle, in degrees. \\
\texttt{kWp} & float32 & Estimated installed capacity, in kWp (rounded to 2 decimals). \\
\texttt{n\_vintages} & Int16, nullable & Number of distinct DeepPVMapper imagery vintages in which the installation was detected. \\
\texttt{first\_seen} & string, nullable & Earliest date the installation is confirmed to exist (oldest matching vintage, or an earlier documented source date). \\
\texttt{last\_seen} & string, nullable & Most recent date the installation is confirmed to exist. \\
\texttt{sources} & string & Comma-separated list of source registry codes that identified this installation (\autoref{tab:registry}). \\
\texttt{frpv\_proba} & float32, nullable & Raw FRPV building probability, when the installation overlaps a FRPV parcel (no threshold applied on this field; floored, not rounded, to 2 decimals so that a very high probability is never displayed as an absolute 1.00). \\
\texttt{false\_positive} & boolean, nullable, tri-state & \texttt{True}: manually confirmed false positive; \texttt{False}: manually confirmed true positive; \texttt{None}: never manually evaluated (the large majority of records -- absence of evaluation must not be read as confirmation). \\
\texttt{false\_positive\_source} & string, nullable & Which manual campaign produced the \texttt{false\_positive} value (\texttt{dpvm\_precision} or \texttt{dpvm\_recall}); \texttt{None} if \texttt{false\_positive} is \texttt{None}. \\
\bottomrule
\end{tabular}
\end{table}

\begin{table}[h]
\centering
\caption{Source registry codes referenced by the \texttt{sources} field.}\label{tab:registry}
\begin{tabular}{c l}
\toprule
\textbf{Code} & \textbf{Source} \\
\midrule
0 & DeepPVMapper \\
1 & FRPV (probability signal only, never a geometry source) \\
2 & OpenStreetMap \\
3 & Manually submitted correction \\
4 & Recall annotation sample (confirmed false negative) \\
6+ & Third-party detection project (one code per project) \\
\bottomrule
\end{tabular}
\end{table}

The \texttt{raw.zip} and \texttt{latest-dpvm.geojson.zip} archives use a narrower, single-source schema (no \texttt{sources}, \texttt{frpv\_proba}, or manual validation fields, since these only exist once sources are cross-referenced) and are documented in full alongside the Zenodo record itself.

\section*{Data Overview}\label{sec:data-overview}

\paragraph{Overall number of installations and descriptive statistics}
\autoref{tab:overview} summarizes the database at a glance; the remainder of this section discusses its source composition and geographic coverage in more detail.

\begin{table}[h]
\centering
\caption{Key descriptive statistics of the OpenPVMapper database (this release).}\label{tab:overview}
\begin{tabular}{l r}
\toprule
\textbf{Statistic} & \textbf{Value} \\
\midrule
Installations & 1,135,850 \\
Estimated installed capacity & 15.0~GWp \\
Estimated PV surface area & 116.1~km$^2$ \\
Mean installed capacity per installation & 13.2~kWp \\
Departments covered (DeepPVMapper / FRPV) & 96/96 -- 95/96 \\
Mainland administrative regions covered & 13/13 \\
Installations corroborated by $\geq$2 independent sources & 430,940 (38\%) \\
\bottomrule
\end{tabular}
\end{table}

\begin{figure}[h]
     \centering
     \includegraphics[width=0.7\linewidth]{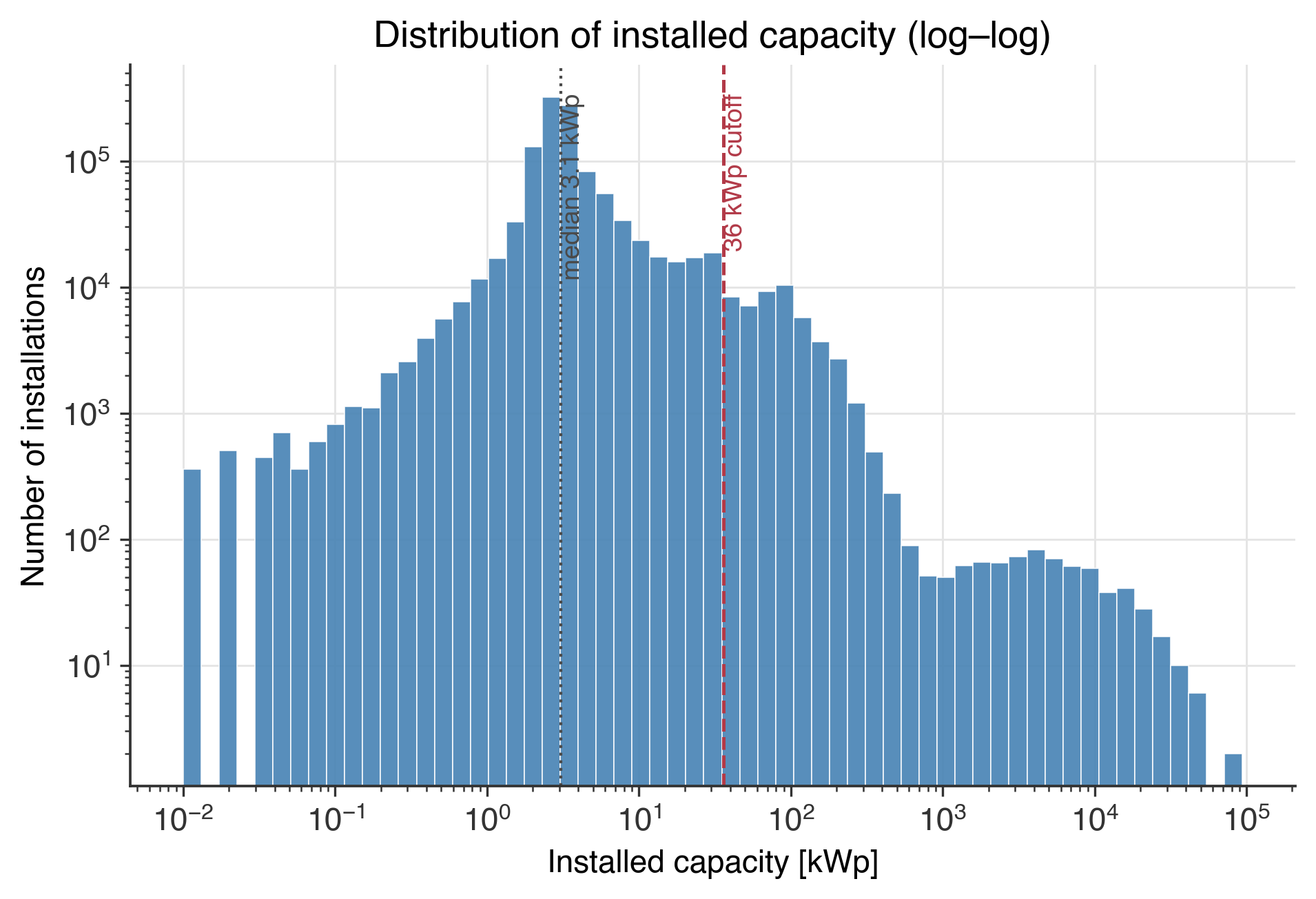}
     \caption{\textbf{Distribution of estimated installed capacity (log-log).} Installed capacity across all 1,135,850 installations, on a log-log scale. The distribution is heavily right-skewed toward the residential/small-commercial segment that dominates the database (mean 13.2~kWp, \autoref{tab:overview}); the tail beyond DeepPVMapper's 36~kWp detection cutoff is populated mostly by OpenStreetMap- and FRPV-contributed installations, consistent with the source/power-class relationship in \autoref{fig:pie-and-power}, plus a small residual population of ground-mounted installations discussed in \emph{Limitations}.}
     \label{fig:kwp-loglog}
\end{figure}

\autoref{fig:kwp-loglog} plots the distribution of installed capacities across the database. The bulk of installations fall in the residential and small-commercial range, consistent with the database's overall composition (\autoref{tab:overview}); a much smaller secondary population above roughly 1~MWp is also visible in the tail and is discussed as a limitation below (\emph{Limitations}) rather than filtered out or ignored here.

\begin{figure}[h]
    \centering
    \includegraphics[width=\linewidth]{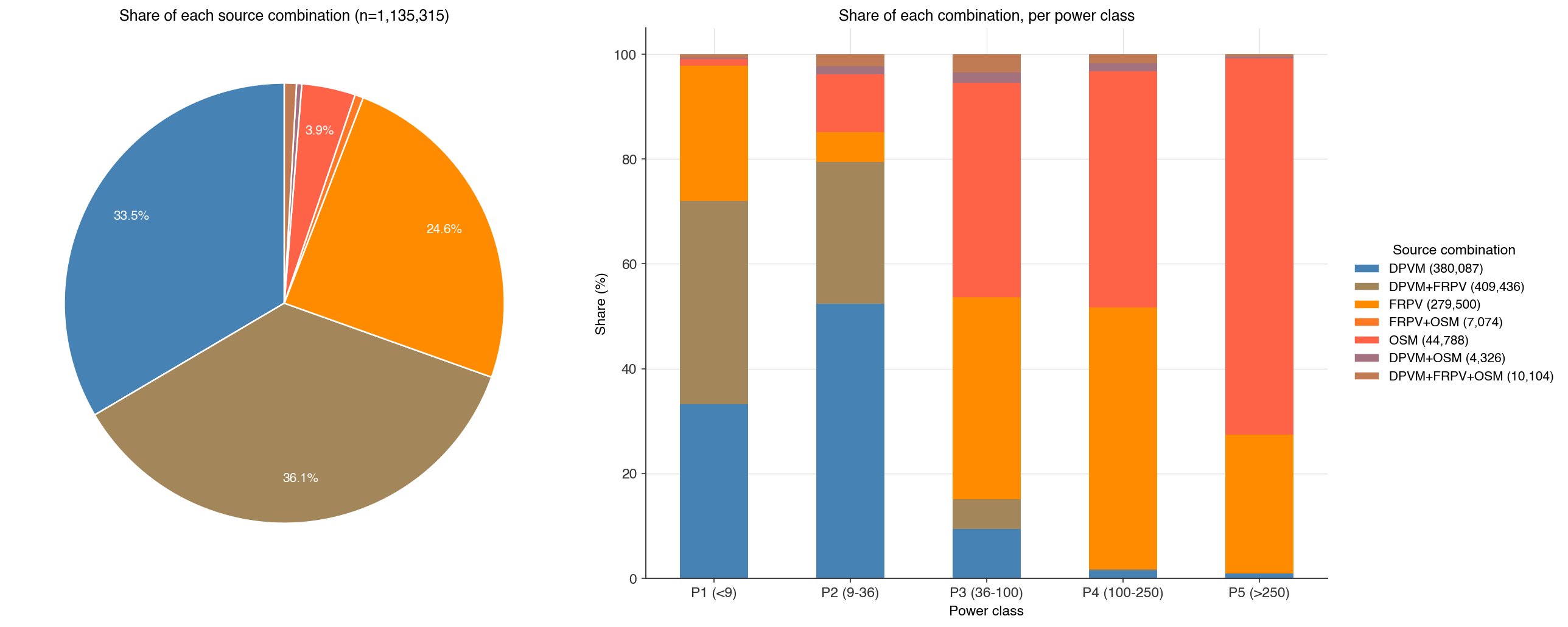}
    \caption{\textbf{Source composition of the database, overall and by power class.} Left: share of installations attributable to each source or combination of sources across the full database ($n=1{,}135{,}315$ installations with a known source combination). Right: the same breakdown within each power class (P1--P5, cf.\ \emph{Technical Validation}).}
    \label{fig:pie-and-power}
\end{figure}

\paragraph{Breakdown by source and power class.}
\autoref{fig:pie-and-power} presents the breakdown of sources overall and by power class. DeepPVMapper and FRPV jointly account for roughly 95\% of the database, with substantial overlap between the two (\autoref{fig:pie-and-power}, left). OpenStreetMap, by contrast, contributes proportionally more large installations: its share rises sharply in the higher power classes (\autoref{fig:pie-and-power}, right), consistent with community mappers being more likely to notice and tag large, visually conspicuous installations than small residential ones -- itself a direct illustration of the coverage gap discussed in \emph{Data Fusion and Enrichment} that motivates combining OSM with automated, independent detection in the first place.

\begin{figure}[htbp]
    \centering
    \begin{subfigure}[b]{0.49\linewidth}
        \centering
        \includegraphics[width=\linewidth]{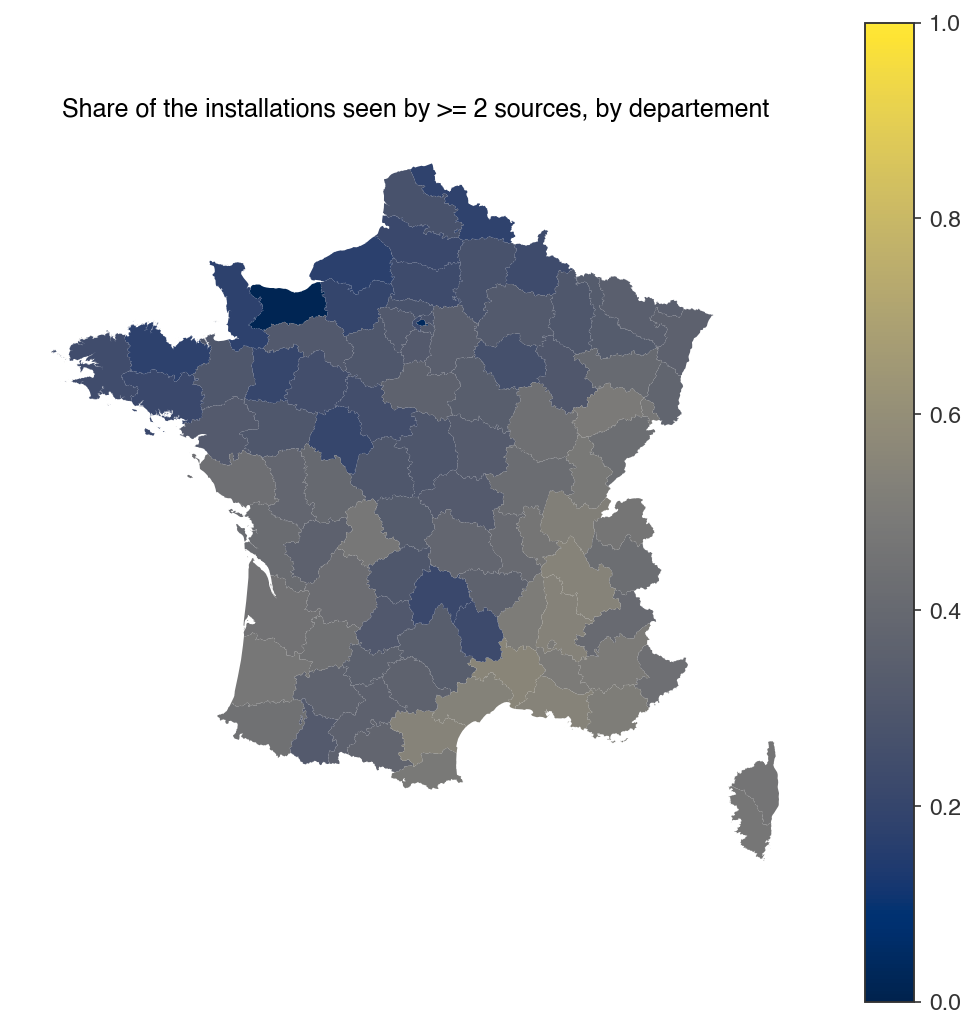}
        \caption{Share of installations confirmed by two or more sources, by department.}
        \label{fig:two_or_more_sources}
    \end{subfigure}
    \hfill
    \begin{subfigure}[b]{0.49\linewidth}
        \centering
        \includegraphics[width=\linewidth]{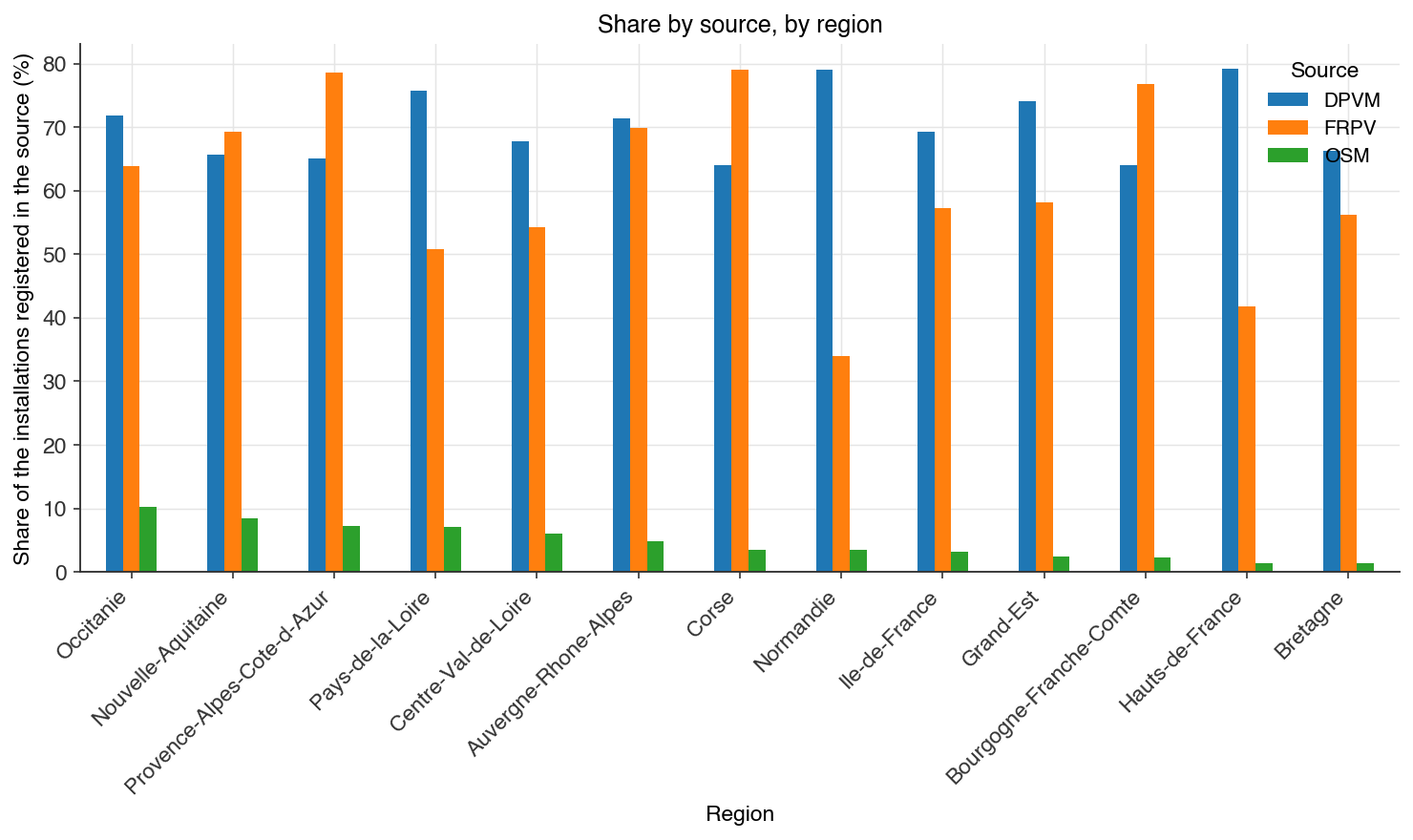}
        \caption{Share of each source's contribution, by region.}
        \label{fig:region_source_coverage}
    \end{subfigure}
    \caption{\textbf{Geographic redundancy of source coverage.} Redundancy is lowest where OpenStreetMap contribution is weakest (north-west) and highest where it is strongest or where FRPV and DeepPVMapper jointly compensate for it, ensuring that most regions retain good multi-source redundancy even where any single source is comparatively weak.}
    \label{fig:global_source_analysis}
\end{figure}

OpenStreetMap's geographic contribution is comparatively small everywhere -- it is never the leading source, by installation count, in any single region (\autoref{fig:region_source_coverage}) -- whereas FRPV and DeepPVMapper alternate in providing the largest regional share, together forming a comparatively stable geographic baseline. OSM's real value lies elsewhere: it disproportionately covers the top of the installed-capacity distribution (\autoref{fig:pie-and-power}, right), corroborating or introducing large installations that the other two sources under-represent (\emph{Limitations} returns to this point). This complementarity translates into good geographic redundancy overall: most departments have a substantial share of their installations confirmed by two or more independent sources, with the lowest redundancy concentrated in a handful of north-western departments where OpenStreetMap's contribution is weakest (\autoref{fig:two_or_more_sources}).

The national database contains 1,135,850 installations after multi-source aggregation and deduplication (8,659 duplicate pairs, detected across sources by geometric overlap or centroid proximity, were merged; see \texttt{dedup\_audit.csv} for the full merge log).
Together, these installations total approximately 15.0~GWp of estimated installed capacity (mean 13.2~kWp per installation) and 116.1~km$^2$ of estimated PV surface area.

Installations are unevenly distributed across source combinations: 36.1\% are confirmed by both DeepPVMapper and FRPV, 33.5\% by DeepPVMapper alone, 24.6\% by FRPV alone, 3.9\% by OpenStreetMap alone, and 0.9\% by all three sources simultaneously -- 430,940 installations (38\% of the database) are corroborated by two or more independent sources (\emph{Technical Validation}, below, quantifies the precision gain this corroboration provides).

\autoref{tab:regions} reports the number of installations per region. Coverage is relatively even for DPVM, which is deployed nationwide, while FRPV (95/96 departments at this release) and OpenStreetMap (contribution-dependent) contribute unevenly across regions -- visible here as regional differences that partly reflect differences in source coverage rather than differences in the underlying PV deployment alone.

\begin{table}[h]
\centering
\caption{Installations per region.}\label{tab:regions}
\begin{tabular}{l r}
\toprule
\textbf{Region} & \textbf{Installations} \\
\midrule
Occitanie & 193,377 \\
Auvergne-Rh\^one-Alpes & 182,113 \\
Nouvelle-Aquitaine & 162,651 \\
Grand Est & 100,248 \\
Pays de la Loire & 93,885 \\
Provence-Alpes-C\^ote d'Azur & 87,535 \\
Hauts-de-France & 75,508 \\
Bourgogne-Franche-Comt\'e & 50,955 \\
Bretagne & 49,781 \\
\^Ile-de-France & 48,624 \\
Normandie & 45,025 \\
Centre-Val de Loire & 41,869 \\
Corse & 4,278 \\
\midrule
Total & 1,135,849$^\ast$ \\
\bottomrule
\end{tabular}

\footnotesize $^\ast$1 installation (of 1,135,850) could not be assigned to one of the 13 mainland regions (unmapped department code) and is omitted from this breakdown.
\end{table}

\section*{Technical Validation}\label{sec:technical-validation}

\subsection*{FRPV Threshold calibration}

\begin{figure}[h]
    \centering
    \includegraphics[width=0.65\linewidth]{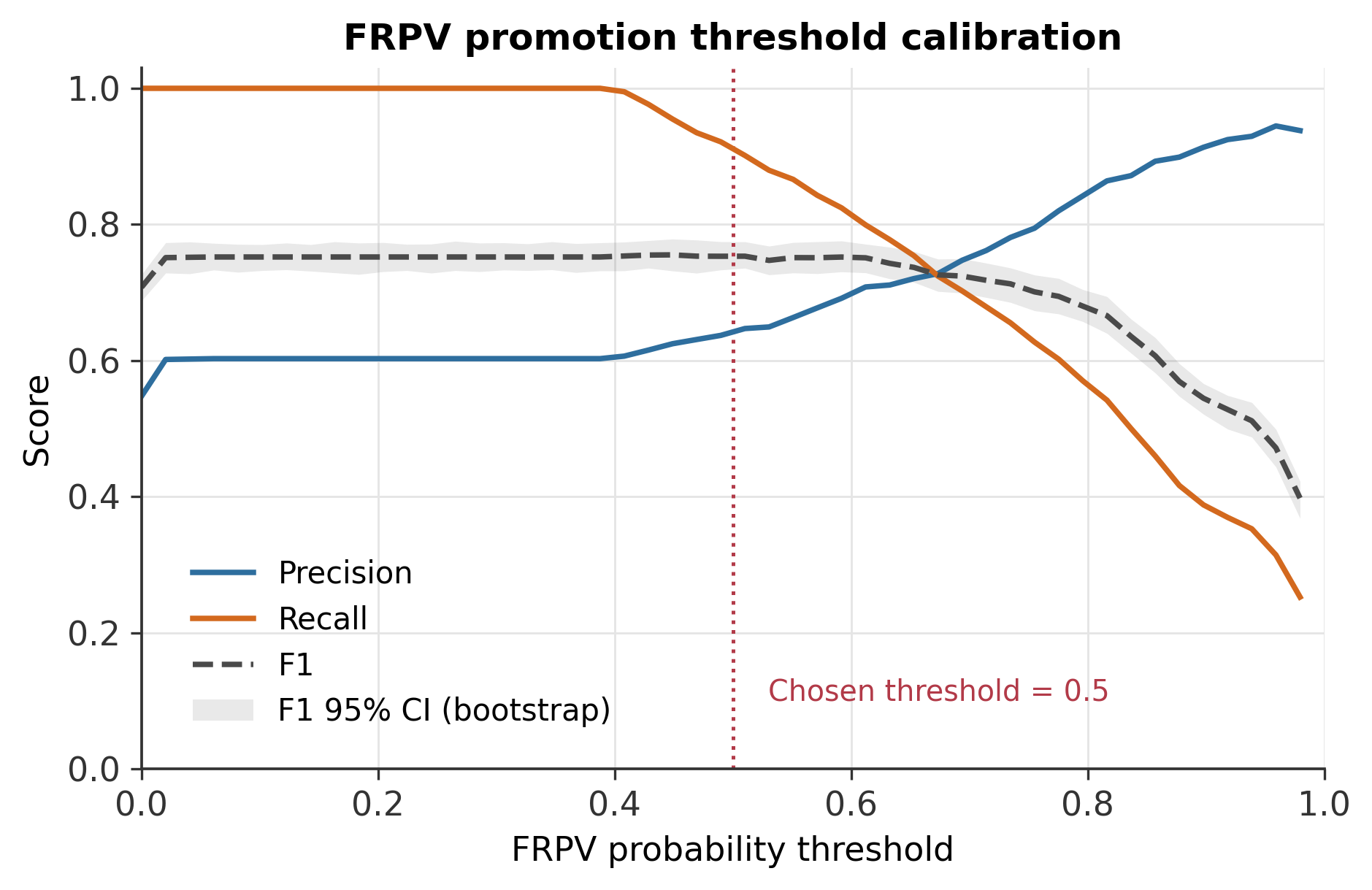}
    \caption{\textbf{FRPV promotion threshold calibration.} Precision, recall, and F1-score (with 95\% bootstrap confidence interval) as a function of the FRPV probability threshold, estimated on the stratified sample of 1,200 manually annotated parcels described above. The retained threshold (0.5, dotted line) sits on a near-flat plateau of the F1 curve, balancing precision and recall rather than maximizing either alone.}
    \label{fig:frpv-threshold}
\end{figure}

FRPV reports a continuous per-building probability rather than a binary detection, so a promotion threshold had to be set before it could contribute candidate installations. As described in \emph{Data Fusion and Enrichment}, this threshold was calibrated on a stratified sample of approximately 1,200 installations, drawn across probability bands to cover the full range of the score distribution and labelled blind to the underlying score, with a precision-recall curve estimated together with bootstrap confidence intervals rather than reported as a single point estimate (\autoref{fig:frpv-threshold}). Two alternative thresholds were considered and discarded: the F1-maximizing threshold (0.45, precision 0.625, recall 0.955) and the threshold reaching 90\% precision (0.90, recall only 0.388, discarding the majority of true positives). The retained threshold, 0.5, lies on a near-flat plateau of the F1 curve (0.752--0.755 between thresholds 0.3 and 0.6) and offers a more balanced precision/recall trade-off than either alternative.

\subsection*{Data quality}

\paragraph{Precision} The enriched database's precision has been evaluated through the manual
review of 1,862 installations, drawn from two independent stratified samples
rather than a single uniform draw: 1,302 installations stratified by the exact
combination of contributing sources (\texttt{sources} field), so as to isolate
the precision gain brought by multi-source corroboration (e.g.\ an installation
confirmed independently by DeepPVMapper and OpenStreetMap versus either source
alone), and 560 installations stratified by installed-capacity class, to check
whether precision varies with installation size independently of source
composition. Annotation was performed with
\href{https://github.com/gabrielkasmi/pv-annotation}{pv-annotation}, an
open-source (MIT license) web-based polygon annotation tool, using ESRI World
Imagery as the verification basemap -- chosen in complement to the IGN orthophoto used for
detection specifically because its can be more recent and thus avoids penalizing
installations recorded in OSM after the IGN imagery was acquired (a stale reference image
would otherwise show a bare roof for a genuine, newer installation). Cases where
the imagery remained inconclusive (occlusion, ambiguous roof state) were tagged
\texttt{unknown} rather than forced into a positive/negative verdict, and
excluded from the precision estimates below rather than counted either way.

Because sample allocation across strata is deliberately disproportionate (smaller or lower-expected-precision strata receive relatively more annotation budget, for comparable margins of error across strata), the pooled, unweighted mean over the annotated sample is not itself a valid estimate of the database's true precision; it must be reweighted by each stratum's true population share within the full database. Doing so for both independent designs yields closely converging estimates -- 74.5\% (source-combination axis) and 73.3\% (power-class axis), a 1.2-point gap -- which we take as a genuine cross-check of the sampling design, since the two axes are stratified independently and share no strata.

As expected, precision rises sharply with corroboration -- 71.5\% for installations confirmed by a single source (n=843), 96.9\% for two sources (n=360), and 98.2\% for three sources (n=57). Nearly all of the gain occurs between one and two sources -- the single result we would highlight above all others in this validation. This corroboration effect is also visible geographically: departments with a higher share of multi-source installations (\autoref{fig:two_or_more_sources}) are, by construction, the departments where this precision gain applies to the largest fraction of the local database. Each asset registered in the dataset indicates its number of sources, enlabling to have a quality label for each installation.

Standalone precision is only $\sim$45\% for DeepPVMapper, versus $\sim$78\% for FRPV alone and $\sim$98\% for OpenStreetMap alone. Consequently, an apparent trend of increasing precision with installation power class is largely a proxy for source composition rather than a direct effect of installation size: small installations are DeepPVMapper-dominated (which has the weakest standalone precision), while large installations are OpenStreetMap-dominated (which has the strongest). This asymmetry, more than corroboration \emph{per se}, is the main driver of the overall precision figures above, and is reported here explicitly rather than left implicit in an aggregate number. It also reinforces the design rationale of \emph{Multi-source acquisition}: OSM's near-perfect standalone precision is exactly what makes it worth treating as the reference geometry whenever it is available, while its limited and uneven coverage is exactly why it cannot be relied on alone.

\paragraph{Recall} Recall is validated indirectly, by comparing the database's connected capacity against Enedis's public grid-connection registry\footnote{\url{https://data.enedis.fr/pages/parc-raccorde/}} as an external proxy, at three successive snapshot dates (Q4 2024, Q2 2025, Q4 2025). \autoref{tab:comparison_enedis} reports, for each power segment, OpenPVMapper's installed capacity against Enedis's connected capacity at each date, along with the resulting coverage ratio.

Coverage decreases mechanically at later snapshot dates: installations connected to the grid after OpenPVMapper's own detection cutoff are, by construction, absent from the database, so comparing against a later Enedis snapshot understates true coverage rather than reflecting a data quality gap. Of the three snapshots, Q2~2025 is likely the most representative comparison point, since it falls closest to the effective acquisition date of the underlying imagery and detections (which extends up to 2025). This timing effect is most visible in the 100--250~kWp segment -- the fastest-growing of the four -- whose apparent coverage falls from 52.3\% (Q4 2024) to 32.2\% (Q4 2025) as newly connected installations accumulate outside the database's own detection window. The sub-36~kWp segment exceeds 100\% coverage at the earliest snapshot (Q4 2024), suggesting that OpenPVMapper may capture some very small installations ahead of, or independently of, their appearance in the Enedis registry.

\begin{table}[h]
\centering
\caption{\textbf{Coverage of grid-connected capacity relative to Enedis registry snapshots.} OpenPVMapper's installed capacity (GWp), by power segment, compared against Enedis's connected-capacity registry at three successive snapshot dates. Coverage (\%) is the ratio of OpenPVMapper's capacity to Enedis's capacity at each date; it decreases over time as a timing artifact (see main text), not as a reflection of database completeness at a fixed point in time.}\label{tab:comparison_enedis}
\resizebox{\textwidth}{!}{%
\begin{tabular}{lrrrrrrr}
\toprule
& \textbf{OpenPVMapper} & \multicolumn{2}{c}{\textbf{Enedis Q4 2024}} & \multicolumn{2}{c}{\textbf{Enedis Q2 2025}} & \multicolumn{2}{c}{\textbf{Enedis Q4 2025}} \\
\textbf{Power segment} & \textbf{(GWp)} & \textbf{GWp} & \textbf{Coverage (\%)} & \textbf{GWp} & \textbf{Coverage (\%)} & \textbf{GWp} & \textbf{Coverage (\%)} \\
\midrule
$<$250~kWp (total)  & 9.07 & 11.54 & 78.6  & 13.43 & 67.5 & 15.28 & 59.4 \\
100--250~kWp        & 2.04 & 3.89  & 52.3  & 5.06  & 40.2 & 6.32  & 32.2 \\
36--100~kWp         & 2.10 & 3.12  & 67.4  & 3.32  & 63.3 & 3.50  & 60.1 \\
$<$36~kWp           & 4.93 & 4.52  & 109.0 & 5.04  & 97.8 & 5.35  & 92.1 \\
\bottomrule
\end{tabular}
}
\end{table}

\subsection*{Accuracy of the data acquisition methods} The accuracy of the underlying detection methods is validated in their respective publications, to which we refer the reader for full detail rather than reproducing their benchmarks here: DeepPVMapper's classification and segmentation performance is reported in \citet{kasmi_enhancing_2024}, and FRPV's building-level detection performance -- including the fine-tuning step used to handle rooftop heterogeneity when deploying the method nationally across France -- is reported in \citet{thebault_comprehensive_2025}. OSM data is filtered on installation-type tags before ingestion, but some edge cases remain despite this filtering (e.g.\ occasional ground-mounted or utility-scale plants tagged in a way that superficially resembles rooftop PV); such cases are corrected as they are identified rather than assumed absent.

\subsection*{Limitations} Three limitations are stated explicitly here rather than left implicit.

First, the boundary between ``rooftop'' and ``ground-mounted'' PV is not a fixed capacity threshold: some legitimate rooftop installations (large industrial or commercial roofs) reach installed capacities in the hundreds of kWp to low MWp range -- the same range as some small ground-mounted plants -- which is why no hard capacity cutoff is applied to distinguish the two. In practice, a small number of ground-mounted installations pass through the pipeline's filters and remain in the released database, visible as the secondary population above roughly 1~MWp in \autoref{fig:kwp-loglog}. This is compounded by the fact that OSM's own installation-type tagging, on which this distinction partly relies, is itself imperfect (\emph{Accuracy of the data sources}, above) and does not always disambiguate the two reliably. Users needing a strict rooftop-only cutoff should filter the released database themselves rather than assume the roof/ground-mounted distinction is fully resolved upstream.

Second, the database does not resolve the persistent ambiguity between an ``installation'' and a ``polygon''. Depending on the source and the roof layout, a single real-world PV installation is sometimes represented by one polygon covering the full array, and sometimes split across several adjacent polygons (e.g.\ one per roof plane or per sub-array) -- and this convention is not necessarily consistent across sources for the same installation. This is documented here rather than resolved: users working with installation-level counts (rather than surface- or capacity-level aggregates, which are more robust to this ambiguity) should be aware that the true number of distinct real-world systems may differ from the row count depending on how multi-polygon arrays are treated.

Third, as noted in \emph{Data Overview}, OpenStreetMap's geographic contribution is a minority everywhere, by installation count -- it is never the leading source in any single region. Its value is not geographic coverage but its disproportionate presence among large installations, where it corroborates or supersedes DeepPVMapper and FRPV far more often than its overall share would suggest (\autoref{fig:pie-and-power}, right).

\section*{Usage Notes}

Because every installation carries its source and, where available, a manual validation flag, users can filter the database according to the level of confidence required by their use case -- for instance restricting to manually validated or multi-source-confirmed installations only. Beyond direct use of the geolocalized installation data (e.g.\ for local or regional rooftop PV potential studies), two use cases motivated this release in particular: cross-checking against grid connection registries to audit their completeness, and feeding validated detections back into OpenStreetMap so that community mapping benefits from this effort rather than duplicating it -- the same interoperability-by-design principle discussed in \emph{Data Fusion and Enrichment} put into practice as a concrete feedback loop.

\section*{Data Availability}\label{sec:data-availability-todo}

The database described in this article is available on Zenodo under a CC-BY license (\texttt{DOI:}~\url{https://doi.org/10.5281/zenodo.21534856}).
Future releases will be published as new versions of the same Zenodo record rather than as separate deposits, so that a single, stable identifier always resolves to the latest release while earlier versions remain individually citable.

\section*{Code Availability}

The source code of the DeepPVMapper detection, segmentation, and characterization pipeline is
available at \url{https://github.com/gabrielkasmi/deeppvmapper/}.

The source code used to construct this database -- combining DeepPVMapper's raw
detections with OpenStreetMap, FRPV, corrections, and the annotation-based
enrichment and deduplication pipeline described in \emph{Data Fusion and Enrichment} and \emph{Uniformization},
together with the code behind the precision/recall validation reported in the
Technical Validation section -- is available at
\url{https://github.com/gabrielkasmi/source-openpvmapper} (archived with a
citable DOI at \url{https://zenodo.org/records/21611312}). Build instructions
and the replication-only files this code needs (model weights, reference
tables, manual-annotation ground truth -- not code, and not part of either
Zenodo data record) are accessible from that repository rather than
duplicated here.

The manual annotation tool used for the precision validation described in the
Technical Validation section, \href{https://github.com/gabrielkasmi/pv-annotation}{pv-annotation},
is released under an MIT license.

\bibliographystyle{unsrtnat}
\bibliography{references}

\appendix


\end{document}